\documentclass[10pt,twocolumn,letterpaper]{article}

\usepackage[pagenumbers]{iccv}
\usepackage[accsupp]{axessibility}
\usepackage{fix-cm}
\usepackage[utf8]{inputenc} 
\usepackage[T1]{fontenc}    
\usepackage{graphicx}
\usepackage{amsmath,bm}
\usepackage{amssymb}
\usepackage{booktabs}
\usepackage{enumitem}
\usepackage{acronym}
\usepackage{balance}
\usepackage{xspace}
\usepackage{setspace}
\usepackage{tabularx,colortbl,multirow,array,makecell}
\usepackage{overpic,wrapfig}
\usepackage{nicefrac}
\usepackage[misc]{ifsym}
\usepackage{siunitx}
\usepackage{pifont}
\definecolor{iccvblue}{rgb}{0.21,0.49,0.74}
\usepackage{hyperref}
\hypersetup{pagebackref,breaklinks,colorlinks,allcolors=iccvblue}
\usepackage[capitalize]{cleveref}
\usepackage{lipsum}

\newcommand{\supp}{\textit{supplementary}\xspace}

\definecolor{gblue}{HTML}{4285F4}
\definecolor{gred}{HTML}{DB4437}
\definecolor{ggreen}{HTML}{0F9D58}
\definecolor{vblue}{HTML}{2993ba}

\definecolor{gbest}{HTML}{FFFFFF}
\definecolor{gsecond}{HTML}{FFFFFF}
\definecolor{gthird}{HTML}{FFFFFF}

\acrodef{git}[GIT]{Gaussian Instance Tracing}
\acrodef{nerf}[NeRF]{Neural Radiance Fields}
\acrodef{2dgs}[2DGS]{2D Gaussian Splatting}
\acrodef{3dgs}[3DGS]{3D Gaussian Splatting}
\acrodef{miou}[mIoU]{mean Intersection over Union}
\acrodef{macc}[mAcc]{mean Accuracy}

\makeatletter
\renewcommand{\paragraph}{%
  \@startsection{paragraph}{4}{\z@}%
  {1ex plus 0.5ex minus 0.2ex} 
  {-1em}                      
  {\normalfont\normalsize\bfseries} 
}
\makeatother

\def\eqref#1{equation~\ref{#1}}

\def\1{\bm{1}}

\def\mW{{\bm{W}}}

\DeclareMathAlphabet{\mathsfit}{\encodingdefault}{\sfdefault}{m}{sl}
\SetMathAlphabet{\mathsfit}{bold}{\encodingdefault}{\sfdefault}{bx}{n}

\def\gG{{\mathcal{G}}}

\def\gM{{\mathcal{M}}}

\def\gR{{\mathcal{R}}}

\def\gU{{\mathcal{U}}}
\def\gV{{\mathcal{V}}}

\def\sR{{\mathbb{R}}}

\def\paperID{1697}
\def\confName{ICCV}
\def\confYear{2025}

\title{Trace3D: Consistent Segmentation Lifting via Gaussian Instance Tracing}
\author{
\begin{tabular}{c}
Hongyu Shen\textsuperscript{1,2}\footnotemark[1] \hspace{0.3em}
Junfeng Ni\textsuperscript{2,3}\footnotemark[1] \hspace{0.3em}
Yixin Chen\textsuperscript{2\ \Letter} \hspace{0.3em}
Weishuo Li\textsuperscript{2} \hspace{0.3em}
Mingtao Pei\textsuperscript{1} \hspace{0.3em}
Siyuan Huang\textsuperscript{2\ \Letter} \\
\small \textsuperscript{1}Beijing Institute of Technology \quad \textsuperscript{2}State Key Laboratory of General Artificial Intelligence, BIGAI\\
\small \textsuperscript{3}Tsinghua University \quad
 \footnotemark[1]\;\;Equal contribution \quad
  \textsuperscript{\Letter}\;\;Corresponding author
    \vspace{3pt}\\
    \href{https://trace-3d.github.io/}{https://trace-3d.github.io/}
    \vspace{-21pt}
\end{tabular}
}

\begin{document}

\twocolumn[{
\renewcommand\twocolumn[1][]{#1}
\maketitle
\begin{center}
    \centering
    \captionsetup{type=figure}
    \includegraphics[width=\linewidth]{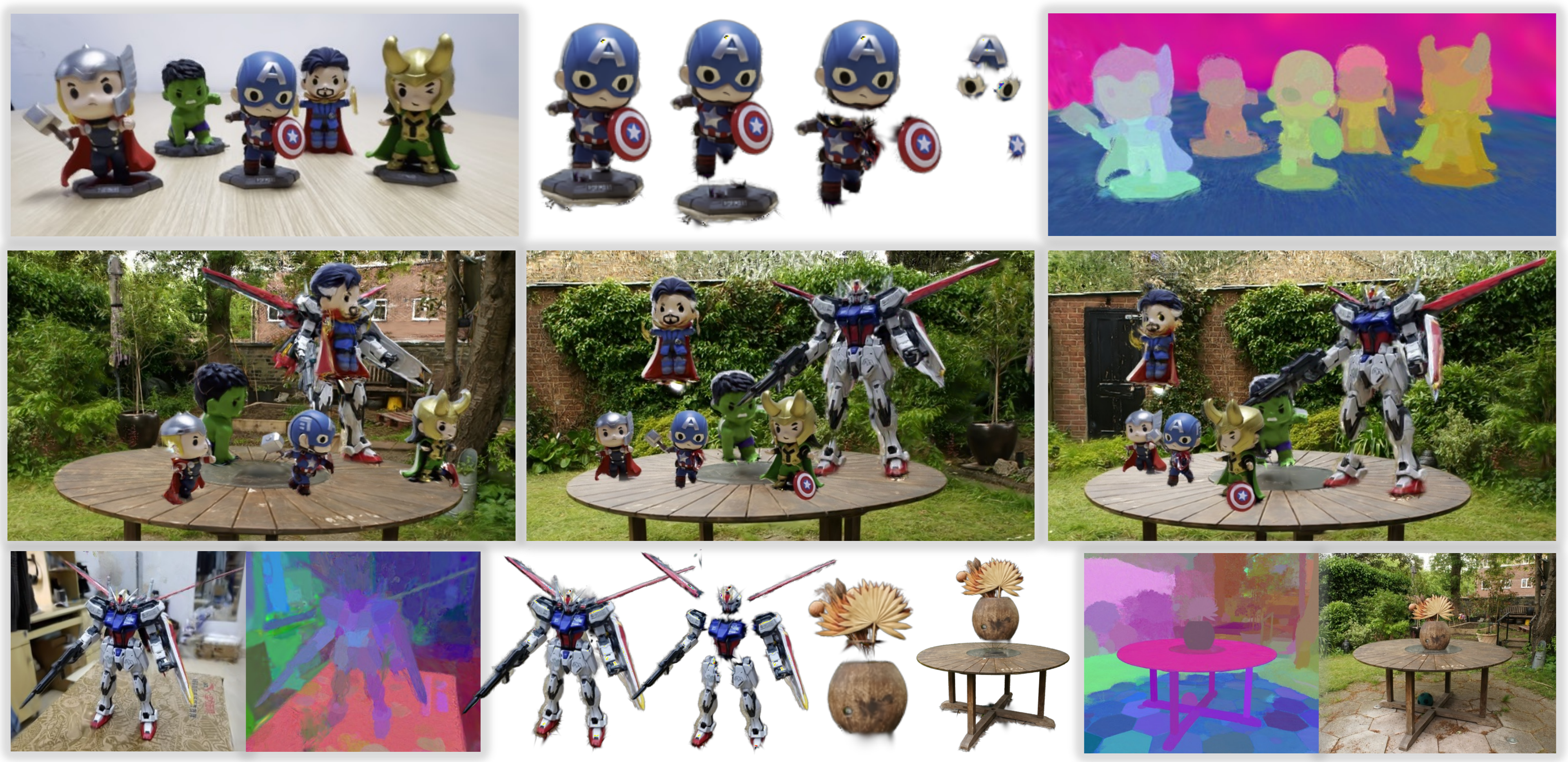}
    \captionof{figure}{
        We propose \textbf{Trace3D} for lifting 2D segmentation to 3D in Gaussian Splatting. We visualize the feature map from PCA and hierarchical segmentations in each scene, facilitating object extraction and scene editing applications, \eg, \textit{Captain America wielding Thor's hammer}. All results are obtained from single reference views after 3D contrastive lifting.
    }
    \label{fig:teaser}
\end{center}
}]

\begin{abstract}
We address the challenge of lifting 2D visual segmentation to 3D in Gaussian Splatting. Existing methods often suffer from inconsistent 2D masks across viewpoints and produce noisy segmentation boundaries as they neglect these semantic cues to refine the learned Gaussians. To overcome this, we introduce \ac{git}, which augments the standard Gaussian representation with an instance weight matrix across input views. Leveraging the inherent consistency of Gaussians in 3D, we use this matrix to identify and correct 2D segmentation inconsistencies. Furthermore, since each Gaussian ideally corresponds to a single object, we propose a \ac{git}-guided adaptive density control mechanism to split and prune ambiguous Gaussians during training, resulting in sharper and more coherent 2D and 3D segmentation boundaries. Experimental results show that our method extracts clean 3D assets and consistently improves 3D segmentation in both online (\eg, self-prompting) and offline (\eg, contrastive lifting) settings, enabling applications such as hierarchical segmentation, object extraction, and scene editing.


\end{abstract}

\section{Introduction}
\label{sec:intro}
3D segmentation~\cite{huang2016point,wang2018depth,tang2020searching,takmaz2023openmask3d} is crucial for holistic scene understanding and underpins various downstream applications, including vision-language reasoning~\cite{3dvista,jia2024sceneverse,huang2024embodied,lu2024manigaussian,yu2025manigaussian++}, embodied AI~\cite{ni2024phyrecon,liu2025building,lu2025dreamart,yu2025metascenes}, and AR/VR~\cite{chen2023ssr,lu2024movis,ni2025dprecon,lu2025taco}. However, advancing 3D segmentation through scaling up remains challenging due to the scarcity of large, densely labeled 3D datasets.  In contrast, the 2D domain has thrived on abundant training data, enabling the development of powerful foundation smodels~\cite{kirillov2023segment,ravi2024sam2} for segmentation.

Building on high-fidelity rendering techniques such as \ac{nerf}~\cite{mildenhall2021nerf} and \ac{3dgs}~\cite{kerbl20233dgs}, a new paradigm emerges that lifts the 2D segmentations from off-the-shelf predictors into 3D. Pioneering work~\cite{zhi2021place,fu2022panoptic,kundu2022panoptic} explores encoding semantic or instance features into the \ac{nerf} representation, while contrastive lifting~\cite{bhalgat2023contrastive} introduces a slow-fast clustering objective to encourage multi-view consistency across frames. More recent works~\cite{qin2024langsplat,gaussian_grouping,gu2025egolifter} adapt these ideas to Gaussian-based representations for faster training and rendering speed, augmenting the Gaussian with additional compact identity encodings or N-channel features.

\begin{figure}[!t]
    \centering
    \includegraphics[width=\linewidth]{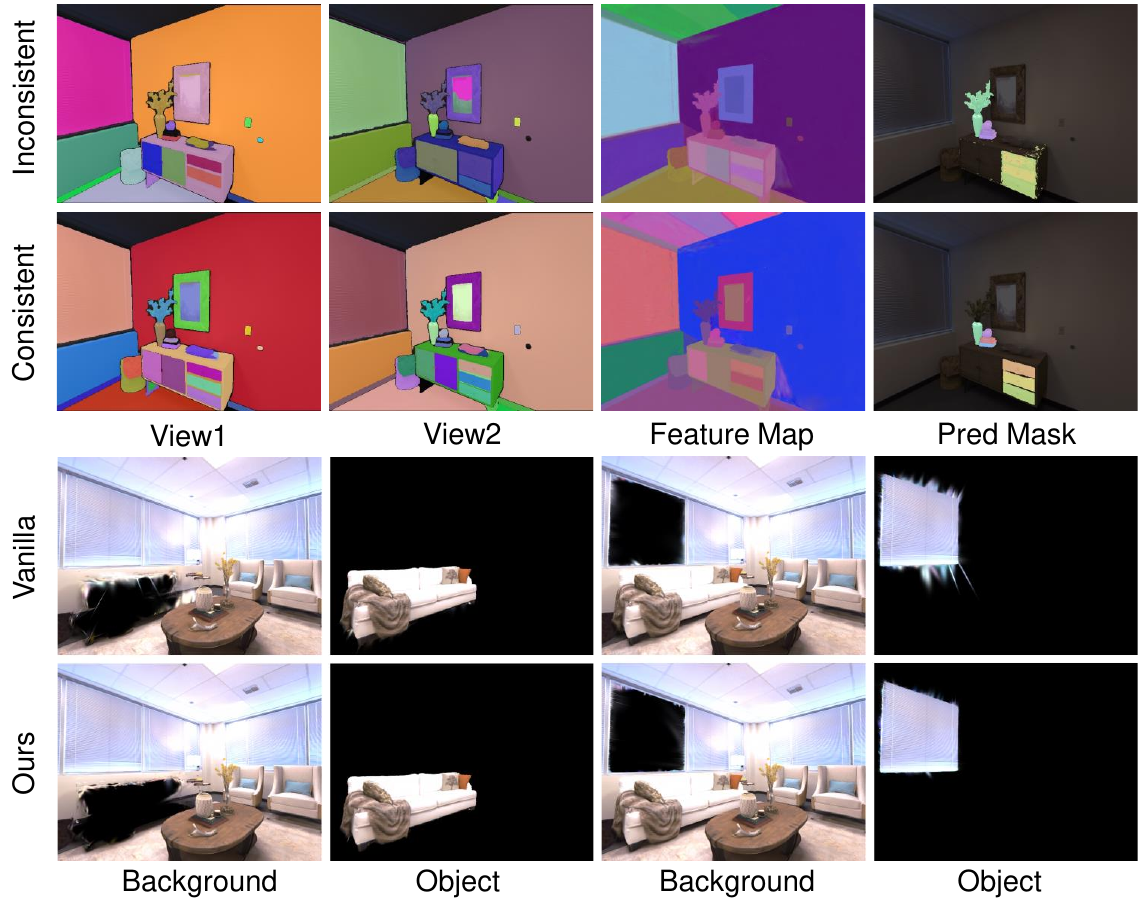}
    \caption{\textbf{Comparisons with existing 3D segmentation lifting.} The inconsistent 2D segmentation masks~\cite{ren2024grounded,kirillov2023segment} introduce significant ambiguity in both feature lifting and mask prediction (top). The resulting boundaries exhibit severe artifacts (bottom), where our method leverages \acf{git} to unify multi-view inconsistent masks and adaptively split or prune ambiguous Gaussians, yielding sharp and clear boundaries.
    }
    \label{fig:intro_problems}
\end{figure}

Despite recent progress, two key challenges persist in Gaussian-based 3D segmentation lifting. First, existing methods, mostly inherited from \acs{nerf}s, \textit{struggle with multiview inconsistencies in 2D segmentation}. Off-the-shelf predictors like SAM~\cite{kirillov2023segment} often produce segmentation masks with varying levels of abstraction for the same objects across different views, as can be seen from \cref{fig:intro_problems}. 
Second,  current approaches \textit{overlook the refinement of the learned Gaussian representation} in this context. While additional 2D segmentation could provide more structured semantic cues to guide and regularize the optimization process, most current approaches rely on fixed Gaussian points obtained from the reconstruction phase. 

Motivated by the well-defined geometric structure of explicit Gaussian representations, we aim to ensure that each Gaussian reliably anchors to the same object across different viewpoints. By leveraging this inherent advantage, we simultaneously address both challenges—enforcing consistent segmentation lifting and refining the Gaussian representation within a unified framework.

Specifically, we introduce \textbf{\acf{git}}, which augments each Gaussian with a set of weights indicating its probability of belonging to different instances across all views. We obtain the weight matrix via a reverse-rasterization step: for each pixel in each view, we track the contributing Gaussians that render it and assign the corresponding probabilities based on the pixel's segmentation label. This process effectively maps each Gaussian to the instance masks it influences, producing a weight matrix spanning all views. Notably, \ac{git} is as fast as forward rendering, making it efficient for both training and inference.

With \ac{git}, we first refine predicted masks into consistent instance maps across viewpoints.
Instead of relying on a single view, we trace their associated Gaussians and perform a majority vote to decide whether two individual masks should be merged. Furthermore, we utilize \ac{git} to identify and resolve ambiguous Gaussians spanning multiple objects by guiding density control and regularizing the Gaussian optimization. By enforcing multi-view consistency and clarifying object boundaries, \ac{git} significantly enhances the quality of 3D segmentation lifting.

Experiments on Replica and NVOS demonstrate that our method enhances the 3D segmentation lifting in both offline (\ie, contrastive lifting) and online (\eg, self-prompting) settings, outperforming all baseline models, including the NeRF-based approaches. Moreover, our method produces clear and sharp 3D segmentation of both foreground objects and background—an essential capability lacking in previous Gaussian-based methods. The extensive ablation study confirms the effectiveness of our proposed component. We also demonstrate that our method works robustly across indoor, outdoor, and desktop scenarios, enabling hierarchical segmentation, object extraction, and various downstream applications in object and scene editing.

Our key contributions are as follows:
\begin{itemize}[leftmargin=*]
    \item We introduce \acf{git}, an efficient and effective tracing operation for 3D segmentation lifting that explicitly associates each Gaussian with its corresponding instances across views via a weight matrix.
    \item Leveraging \ac{git}, we resolve multiview inconsistencies in instance masks and refine ambiguous Gaussians spanning multiple objects through \ac{git}-guided density control during Gaussian optimization.
    \item Our method surpasses all baselines on Replica and NVOS for 3D segmentation. It delivers sharp foreground and background segmentation, supports asset extraction, and generalizes well to in-the-wild videos, with great potential in scene understanding and editing applications.
\end{itemize}

\section{Related Work}
\label{sec:related_work}

\subsection{3D Segmentation by Lifting 2D Mask}
Research on 3D segmentation has explored various representations, including RGB-D images~\cite{wang2018depth}, point clouds~\cite{zhao2021point,huang2016point,qi2017pointnet}, voxel grids~\cite{tang2020searching}, and more.
However, the limited scale of available data has caused 3D segmentation~\cite{Schult23ICRA,takmaz2023openmask3d,ngo2023isbnet,guo2024sam-graph} to lag behind its 2D counterparts~\cite{cheng2021maskformer,kirillov2023segment,ren2024grounded}.
Inspired by the progress in neural rendering, \eg, \acp{nerf}~\cite{mildenhall2021nerf,barron2021mipnerf,barron2023zipnerf}, recent studies~\cite{fu2022panoptic,kundu2022panoptic,lerf2023,wang2022dmnerf,kobayashi2022distilledfeaturefields,engelmann2024opennerf,lyu2024total,yu2024panopticrecon,cen2023segment,ying2024omniseg3d,kim2024garfield,zhang2025monoinstance} propose to tackle 3D segmentation by lifting 2D mask predictions from 2D foundation models, \eg, SAM~\cite{kirillov2023segment}, into 3D space.
These methods generally fall into the \textit{offline} and \textit{online} settings. Offline methods~\cite{siddiqui2023panoptic,bhalgat2023contrastive,gu2025egolifter} rely on contrastive learning~\cite{hadsell2006dimensionality,chen2020simple} to train a feature field, pulling together the features from the same object while pushing apart those from different objects. 
Online methods, like SA3D~\cite{cen2023segment} and FlashSplat~\cite{shen2025flashsplat} perform iterative cross-view self-prompting to generate SAM masks on novel views and refine the 3D mask of the target object.  
However, both settings suffer from inconsistent 2D masks across viewpoints and hierarchy, as illustrated in \cref{fig:intro_problems}.
In contrastive lifting, these inconsistencies introduce ambiguity, as the same 3D features may be supervised to render the same mask in one view and diverge in another, especially when considering hierarchy like OmniSeg3D~\cite{ying2024omniseg3d} and GARField~\cite{kim2024garfield}. Likewise in self-prompting, they cause ambiguity in the 3D masks which are optimized via majority voting across input views.
In this paper, we address this fundamental problem by tracing the instance weight of all Gaussians through \ac{git} and resolving inconsistent masks by leveraging the consistent nature of Gaussians in 3D.


\subsection{3D Segmentation on Gaussian Splatting}
Gaussian Splatting has emerged as an efficient neural rendering approach, enabling the reconstruction of 3D scenes from 2D images through learning
explicit 2D~\cite{huang20242d} or 3D~\cite{kerbl20233dgs} Gaussians. Recently, several works ~\cite{cen2023sags,chen2023gaussianeditor,wu2024opengaussian,contrastive_gaussians_2024,guo2024semantic,zhou2024feature,shen2025flashsplat} have extended this framework to lift 2D masks to 3D segmentation; see comparison in \cref{tab:model_compare}.
However, these methods primarily benefit from the speed advantages of Gaussain Splatting, \ie, fast training and rendering, without fully exploiting its unique characteristics or addressing its special problems.
For example, GaussianCut~\cite{jain2024gaussiancut}, CoSSegGaussians~\cite{dou2024cosseggaussians} and GaussianGrouping~\cite{gaussian_grouping} use video object trackers to associate masks across different views, but still suffer from lost tracklets and inconsistent masks across different hierarchy. In contrast, we achieve consistent masks by tracing the instance weights for Gaussians and ensure their consistency in 3D.
While tracing via reverse rendering is utilized in GaussianEditor~\cite{chen2024gaussianeditor} for semantic editing, our core contribution lies in maintaining a global instance weight for every Gaussian across all input views, as opposed to Gaga~\cite{lyu2024gaga}, which groups the 3D Gaussians by considering the masks sequentially with a memory bank.

Furthermore, since a 3D scene is essentially a collection of unstructured Gaussians, grouping them based on object semantics inevitably results in ambiguous Gaussians, \ie, those associated with multiple objects. This ambiguity often manifests noticeable artifacts at object boundaries.
Previous works like GaussianGrouping~\cite{gaussian_grouping} or Egolifter~\cite{gu2025egolifter} simply ignore these ambiguous Gaussians~\cite{gaussian_grouping,gu2025egolifter}; FlashSplat~\cite{shen2025flashsplat} and SAGS~\cite{cen2023sags} filter them out, and SAGD~\cite{hu2024sagd} decompose the boundary Gaussians as a post-processing step, causing critical detail loss on the foreground and background boundaries. Our method handles them more robustly by adaptively optimizing the Gaussians with a \ac{git}-guided density control, \ie, splitting and pruning the ambiguous Gaussian. This yields more accurate 3D object segmentation with fewer artifacts.

\begin{table}[ht]
    \centering
    \small
    \caption{\textbf{Method comparison in 3D segmentation lifting.} ``SP.'' and ``Contr.'' denote Self-Prompting and Contrastive Lifting. Our method supports ``Multi-Object'' 3D segmentation instead of ``Binary'' foreground, achieves consistent 2D masks without video tracker, and offers density control for ambiguous Gaussains.}
    \resizebox{\linewidth}{!}{
    \begin{tabular}{lcccccc}
        \toprule
        \multirow{2}{*}{Method}                   & \multirow{2}{*}{Rep.} & \multirow{2}{*}{Lifting} & Video & 3D & Consistent & Amb. Gauss.\\
                                                  &      &         & Tracker       & Segmentation  & 2D Mask         & Control\\
        \midrule
        OmniSeg3D~\cite{ying2024omniseg3d}        & NeRF & Contr.  & $\times$      & Multi-Object  & $\times$        & -   \\
        SA3D~\cite{cen2023segment}                & NeRF & SP.     & $\times$      & Binary        & $\times$        & -  \\
        SA3D-GS~\cite{cen2023sags}                & GS & SP.     & $\times$      & Binary        & $\times$        & $\times$   \\
        Egolifter~\cite{gu2025egolifter}          & GS & Contr.  & $\times$      & Multi-Object  & $\times$        & $\times$   \\
        FlashSplat~\cite{shen2025flashsplat}      & GS & SP.     & $\times$      & Binary        & $\times$        & $\times$   \\
        GaussianCut~\cite{jain2024gaussiancut}    & GS & SP.     &  $\checkmark$ & Binary        & $\checkmark$    & $\times$   \\
        Ours                                      & GS & Both    & $\times$      & Multi-Object  & $\checkmark$    & $\checkmark$   \\
        \bottomrule
    \end{tabular}
    }
    \label{tab:model_compare}
\end{table}

\section{Method}
\label{sec:method}

\begin{figure*}[!t]
    \centering
    \includegraphics[width=\linewidth]{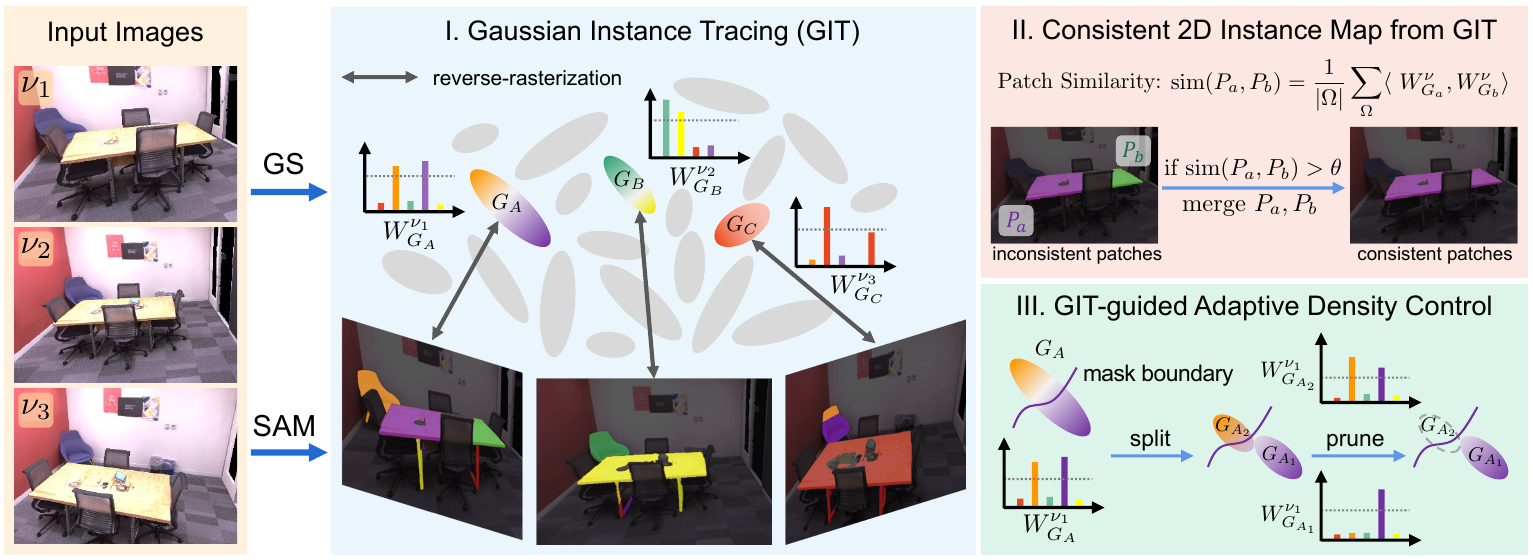}
    \caption{\textbf{Overview of Trace3D.} Given input images and their SAM masks, we employ \ac{git} to compute the instance weight matrix $W$ for all Gaussians (\cref{sec:git}), which is then utilized to merge inconsistent patches based on their similarity (\cref{sec:consistent_mask}). We also propose \ac{git}-guided density control to split and prune the ambiguous Gaussians (\cref{sec:density_control}). Collectively, they contribute to more robust 3D segmentation lifting (\cref{sec:lifting}) in both online and offline settings. For clarity, only a subset of SAM patches is visualized.}
    \label{fig:method}
\end{figure*}

\subsection{Preliminary: \texorpdfstring{\acf{2dgs}}{}}
\label{subsec:2dgs}
\acs{2dgs}~\cite{huang20242d} is proposed and shown to be more suitable for recovering accurate surfaces as it collapses the 3D volume of \ac{3dgs}~\cite{kerbl20233dgs} into a set of 2D oriented planar Gaussian disks, each characterized by a center point $\mathbf{p}_k$, tangential vectors $\mathbf{t}_u$ and $\mathbf{t}_v$, and two scaling factors $s_u$ and $s_v$ controlling the variance. 
For the point $\mathbf{u} = (u, v)$ in $uv$ space, its 2D Gaussian value can be evaluated using the standard Gaussian function:
\begin{equation}
    g(\mathbf{u}) = \exp \left(-\frac{u^{2}+v^{2}}{2}\right).
\end{equation}
Each 2D Gaussian has opacity $\alpha$ and view-dependent appearance \textbf{c} with spherical harmonics. For rasterization, Gaussians are sorted according to their depth and composed into an image with front-to-back alpha blending:
\begin{equation}
    \mathbf {c}(\mathbf {x}) = \sum_{i=1} \mathbf {c}_i \alpha_i g_i(\mathbf {u}(\mathbf {x})) \prod_{j=1}^{i-1} \left(1 - \alpha_j g_j(\mathbf {u}(\mathbf {x}))\right),
\end{equation}
where $\mathbf {u}(\mathbf{x})$ represents the $uv$ intersection position of the ray from the camera origin to pixel $\mathbf{x}$ passing 2D Gaussians.

\subsection{\texorpdfstring{\acf{git}}{}}
\label{sec:git}
From a set of input images $\{I_i\}_{i=1}^{L}$ with known camera parameters, we reconstruct a \ac{2dgs} scene parameterized by $\gG = \{G_i\}_{i=1}^{N}$. For each image $I^\nu, \nu\in\{1, \dots, L\}$, we use a 2D segmentation model, \eg, SAM~\cite{kirillov2023segment}, to obtain class-agnostic binary masks $\gM^\nu = \{M_j^\nu | j = 1, \cdots , m^\nu\}$. 

We then form a single \textit{instance map} by overlapping all masks and treating each overlapping region as a disjoint instance patch,
\ie, $\gM^\nu \rightarrow \{P^\nu_j | j = 1, \dots, p^\nu\}$. All subsequent operations in this paper are performed on these primitive-level patches, which is different from overlaying masks one-by-one in Egolifter~\cite{gu2025egolifter} or storing the hierarchy for each image like OmniSeg3D~\cite{ying2024omniseg3d}. More detailed comparisons are provided in \supp.

Note that both the binary masks and instance maps can be inconsistent across different views and mixed in hierarchies, as shown in \cref{fig:method}. To address these issues, we propose \ac{git}, which first augments each Gaussian $G_i$ with a weight matrix $\mW_i \in \sR^{T \times L}$, where $T=\max_\nu p^\nu$ is the maximum number of disjoint instance patches in any single view. $\mW_i^\nu$ represents the instance distribution with each entry $\mW_{i,j}^\nu$ for the probability that the $i$-th Gaussian $G_i$ belongs to the $j$-th instance under view $\nu$. We obtain the weight matrix via a \textit{reverse-rasterization} step: for each pixel in view $\nu$, we identify the Gaussians responsible for splatting onto that pixel. We then attribute the pixel's instance label to these Gaussians in proportion to their contribution. This process traces each Gaussian back to the instance patches it influences: summing and normalizing over these per-pixel attributions yields the probability distribution $\mW_i^\nu$ of each Gaussian's association with different instances in view $\nu$. Note that the \ac{git} operation is computed in parallel similar to the forward rendering, making it efficient and flexible to use during training and inference.

\subsection{Consistent 2D Instance Map from \texorpdfstring{\acs{git}}{}}
\label{sec:consistent_mask}

We aim to merge the patches into a refined set of instance maps that are consistent across views and hierarchy, which serve as the basis for Gaussian optimization and robust 3D segmentation lifting in subsequent stages. Consider two patches $P_a^\nu, P_b^\nu$ from an initial instance map under view $\nu$. We cannot determine whether they should be merged based solely on the current view. Instead, we choose to believe the majority vote of whether their corresponding Gaussians belong to the same patch from other views. 

Specifically, we first trace the Gaussian set $\gG_a, \gG_b$ from the patch $P_a^\nu, P_b^\nu$, which are determined from the weights matrix $W$ by collecting all Gaussians that have non-zero probability of falling into $P_a^\nu$ or $P_b^\nu$, respectively. We then find the Gaussians from both sets that are co-visibile from the views, forming the set of $\Omega = \{(G_a, G_b, \nu) | G_a\in\gG_a, G_b\in\gG_b, \nu\in\{1, \dots, L\}\}$. We define the similarity of the two patches $P_a^\nu, P_b^\nu$ by:
\begin{align}
  \mathrm{sim}(P_a,P_b)&=\frac{1}{|\Omega|} \sum_{(G_a,G_b,\nu)\in\Omega} \langle\ W_{G_a}^\nu,W_{G_b}^\nu\rangle,
  \label{eq:gsim}
\end{align}
where the $\langle\cdot,\cdot\rangle$ measure the similarity between the Gaussians' instance probability distributions $W_{G_a}^\nu, W_{G_b}^\nu \in \sR^T$ with inner product.
Patches with a similarity score exceeding a pre-determined threshold $\theta=0.5$ will be merged. By iterating this procedure for all pairs of patches within views, we obtain a final set of 2D instance maps consistent across multi-view observations and hierarchical configurations.

\subsection{\texorpdfstring{\acs{git}}{}-guided Adaptive Density Control}
\label{sec:density_control}
One major challenge in training the Gaussian Scene $\gG$ from multi-view RGB images, without explicitly considering object boundaries, is the emergence of \textbf{ambiguous Gaussians} that overlap between different objects. These ambiguous Gaussians complicate the alignment between 3D segmentation and the underlying Gaussian representation, hindering the quality of 3D segmentation lifting and further downstream tasks. Thus, we propose to use \acs{git} to guide density control and regularize the Gaussian optimization.

\subsubsection{Ambiguous Gaussian Detection}
Given a trained Gaussian scene $\gG$, we use our consistent patch maps to get a refined weight matrix $W$ for all Gaussians. For each Gaussian $G_i$, its ideal patch probability distribution $W_i^\mathbf{v}$ should be nearly one-hot on visible views and all-zero on non-visible views. Thus, we define \textbf{ambiguous Gaussians} to have significant non-zero weights across two or more objects.
Denoting the set of visible view of $\gG_i$ as $\gV_i$, its ambiguity score $As_i$ is computed as:
\begin{equation}
As_i=\frac{1}{|\gV_i|}\sum_{\nu\in \gV_i }{\mathbb{I}\left(\max_j(W_{i,j}^{\mathbf{v}}
)<\gamma \right )},
\end{equation}
where $\mathbb{I}$ is the indicator function. Gaussians with an ambiguity score higher than $\theta_{As}$ are detected as ambiguous. We set $\gamma=0.8$ and $\theta_{As}=0.5$ in the experiments.

\subsubsection{Adaptive Density Control}
Directly removing the ambiguous Gaussians in post-processing, as in SA3D-GS~\cite{cen2023sags} or FlashSplat~\cite{shen2025flashsplat}, may result in surface artifacts on the objects and background. Instead, we propose to perform an online Gaussian optimization with \acs{git}-guided density control after detecting these Gaussians. More specifically, we replace the strategy in \ac{3dgs} and split each ambiguous Gaussian into two smaller ones by dividing their scale by 2.0. Their new positions are initialized by using the original Gaussian as a PDF for sampling. Gaussians that remain ambiguous after splitting are removed. The process is repeated every 1000 iterations while retraining the entire Gaussian scene. This refines boundary regions, mitigates ambiguity, and improves overall segmentation quality.

\subsection{3D Segmentation Lifting}
\label{sec:lifting}
Our method both enforces multi-view instance map consistency and clarifies object boundaries, and thus benefits 3D segmentation through either offline contrastive lifting~\cite{bhalgat2023contrastive} or online self-prompting~\cite{shen2025flashsplat,cen2023sags}.
\paragraph{Contrastive Lifting}
3D segmentation with contrastive lifting~\cite{ying2024omniseg3d,gu2025egolifter} augments each Gaussian $G_i$ with an extra feature vector $f_i\in \gR^d$, which supports rendering to a 2D feature map $F\in\gR^{H\times W\times d}$ with differentiable rasterization. The contrastive loss is computed over a set of sampled pixels $\gU$ on the image:
\begingroup
\small
\begin{equation}
L_{contr} = -\frac{1}{|\gU|}\sum_{u\in\gU}log\frac{\sum_{u'\in\gU^+} \text{exp}(\text{sim}(F[u],F[u'];\tau)}{\sum_{u'\in\gU} \text{exp}(\text{sim}(F[u],F[u'];\tau)},
\end{equation}
\endgroup
where $\gU^+$ is the set of pixels belonging to the same instance as $u$. The similarity function is $\text{sim}(f_1,f_2;\tau)=\text{exp}(-\tau||f_1-f_2||^2)$ with $\tau$ as a temperature.
We use the consistent instance maps from \cref{sec:consistent_mask} and the refined Gaussian set $\gG$ from \cref{sec:density_control} to train the features. During inference, query features from the object of interest in the reference view are used to either obtain the target set of Gaussians in 3D for object extraction or derive the 2D segmentation mask on novel views by directly comparing them against the rendered 2D features.

Building on our consistent 2D instance map at the patch level, we further devise an iterative query point-finding strategy that adaptively locates suitable query features and thresholds in accordance with the reference mask, yielding precise segmentations in 3D or novel views. This enables 3D segmentation and object extraction at various hierarchies or user-defined combinations in the experiments.

\paragraph{Self-Prompting}
Following the online setting of FlashSplat~\cite{shen2025flashsplat} and SA3D~\cite{cen2023segment}, the self-prompting workflow begins with user-provided point prompts on a single reference view, which SAM uses to generate the initial mask. Then, we utilize \ac{git} to extract new point prompts on novel views and use SAM to produce additional masks. By iterating this procedure across more views, we obtain a consistent and complete 3D Gaussian set from our weight matrix $\mW$.  Finally, these Gaussains are rendered onto the test view to produce the corresponding 2D instance mask.

\subsection{Implementation details}
 We use the default settings of \ac{2dgs}~\cite{huang20242d} to obtain the initial Gaussian set. For contrastive lifting, the instance feature dimension $d$ is set to be 16, optimized by an Adam optimizer with an initial learning rate of $1\times 10^{-5}$, and temperature of 0.01. All experiments were conducted on a single RTX 4090 GPU. 
 
 
\begin{figure*}[t!]
    \centering
    \includegraphics[width=\linewidth]{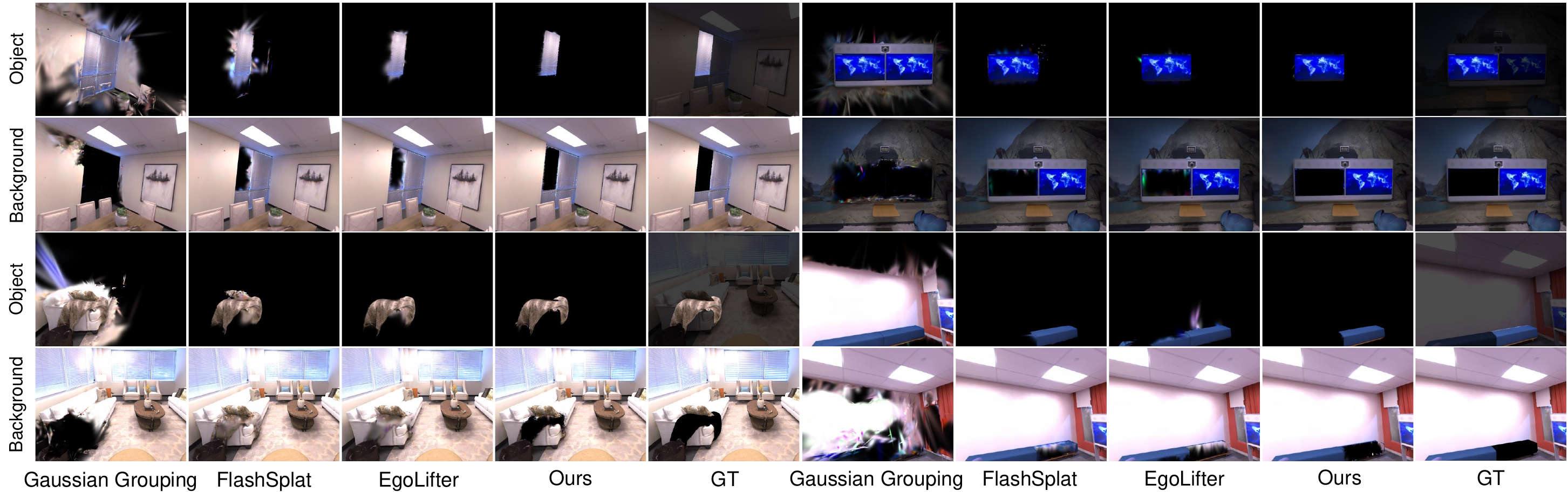}
    \caption{\textbf{Qualitative results of 3D object extraction on Replica.} Our method yields cleaner and sharper object boundaries on both the foreground and background Gaussians.}
    \label{fig:qual_object}
\end{figure*}

\begin{table*}[t!]
    \centering
    \small
    \caption{\textbf{Quantitative results of 3D object extraction on Replica.} The mIoU for each scene is listed, together with the mean mIoU and PSNR for all scenes. The detailed PSNR reports are in \supp.}
    \resizebox{\linewidth}{!}{
    \begin{tabular}{l|ccccccccc|c}
        \toprule
         Method & office0 & office1 & office2 & office3 & office4 & room0 & room1 & room2 & \textit{avg.} mIoU &\textit{avg.} PSNR \\
         \midrule
         Gaussain Grouping~\cite{gaussian_grouping}&23.7&45.9&25.5&30.6&30.2&22.5&38.5& 20.1&29.6&13.4\\
         FlashSplat~\cite{shen2025flashsplat}&47.5&45.9&39.6&36.9&27.5&40.6&39.8&36.6&39.3&16.9\\
         Egolifter~\cite{gu2025egolifter}&67.4&59.6&48.9&54.8&59.4&50.7&53.7&50.1&55.6&20.1\\
         Ours&\textbf{80.7}&\textbf{76.0}&\textbf{66.1}&\textbf{69.8}&\textbf{71.7}&\textbf{67.0}&\textbf{72.0}&\textbf{73.4}&\textbf{72.1}& \textbf{22.6}\\
        \bottomrule
    \end{tabular}
    }
    \label{tab:gaussian_seg}
\end{table*}
\section{Experiment}
We assess the efficacy of our approach by evaluating both the quality of 3D object extraction and the accuracy of novel view 2D instance segmentation. The failure cases and discussion of limitations are in \supp.

\subsection{3D Object Extraction}
\label{sec:3d_obj_extract}

Previous work~\cite{gaussian_grouping,shen2025flashsplat,gu2025egolifter} only evaluates the final rendered instance map, where all Gaussians contribute through alpha blending. We adopt this evaluation protocol in \cref{sec:2d_ins}, but note that it can obscure artifact Gaussians with low alpha value yet large scale as they are averaged out during the alpha-blending process. In contrast, our goal of 3D segmentation is to produce artifact-free 3D assets. Therefore, we follow prior work~\cite{lee2024rethinking,chacko2025liftinggaussian} and directly evaluate 3D object extraction on the Gaussian segmentations.

\paragraph{Settings}
We evaluate 3D object extraction on the 360$^{\circ}$ captured Replica~\cite{replica19arxiv} dataset and compare our method with Gaussian Splatting-based baselines, including Gaussian Grouping~\cite{gaussian_grouping}, FlashSplat~\cite{shen2025flashsplat} and Egolifter~\cite{gu2025egolifter}. Each method extracts the 3D Gaussian segmentations from the query image and mask of the target object. For our method, we first lift the 2D segmentation masks to 3D by contrastive lifting as in ~\cref{sec:lifting}, and get 2D segmentation masks on novel views. The 3D Gaussians are traced from these multi-view masks. For evaluation, the 3D Gaussians are rendered at novel views, with the mask quality assessed with \ac{miou} and color fidelity with PSNR. We evaluate the objects with GT segmentation annotation from the Replica dataset, with the full list detailed in \supp following SA3D~\cite{cen2023sags}.

\paragraph{Results}

As shown in ~\cref{tab:gaussian_seg} and ~\cref{fig:qual_object}, our method delivers significantly more accurate and cleaner 3D object extraction with fewer artifacts. 
These improvements firstly stem from our consistent 2D instance maps, which support featuring learning in contrastive lifting by eliminating the multi-view mask ambiguity present in SAM~\cite{kirillov2023segment}. Additionally, our \acs{git}-guided adaptive density control, which efficiently detects and separates ambiguous Gaussians, contributes more significantly to the boundaries, leading to sharp boundaries on both the desired objects and background. 
As illustrated in ~\cref{fig:qual_object}, when extracting the Gaussian set of the target object, Gaussian Grouping~\cite{gaussian_grouping} struggles with noisy mask predictions, \eg, mixing parts of both the \textit{sofa} and \textit{blanket} when only \textit{blanket} is desired. While FlashSplat~\cite{shen2025flashsplat} and EgoLifter~\cite{gu2025egolifter} successfully extract correct instances, they leave noticeable artifacts in both the extracted foreground Gaussians and the remaining background. 
In contrast, our method not only accurately extracts the target object but also minimizes artifacts, leading to superior mIoU and PSNR compared to all baselines.

\subsection{Novel View 2D Instance Segmentation}
\label{sec:2d_ins}

\paragraph{Settings} 
Following the common practice of prior work~\cite{cen2023segment,ren-nvos,ying2024omniseg3d,gu2025egolifter}, we also evaluate novel view 2D instance segmentation on the Replica~\cite{replica19arxiv} and NVOS~\cite{ren-nvos} benchmarks using \ac{miou} and mAcc as metrics. 

\begin{figure*}[!t]
    \centering
    \includegraphics[width=\linewidth]{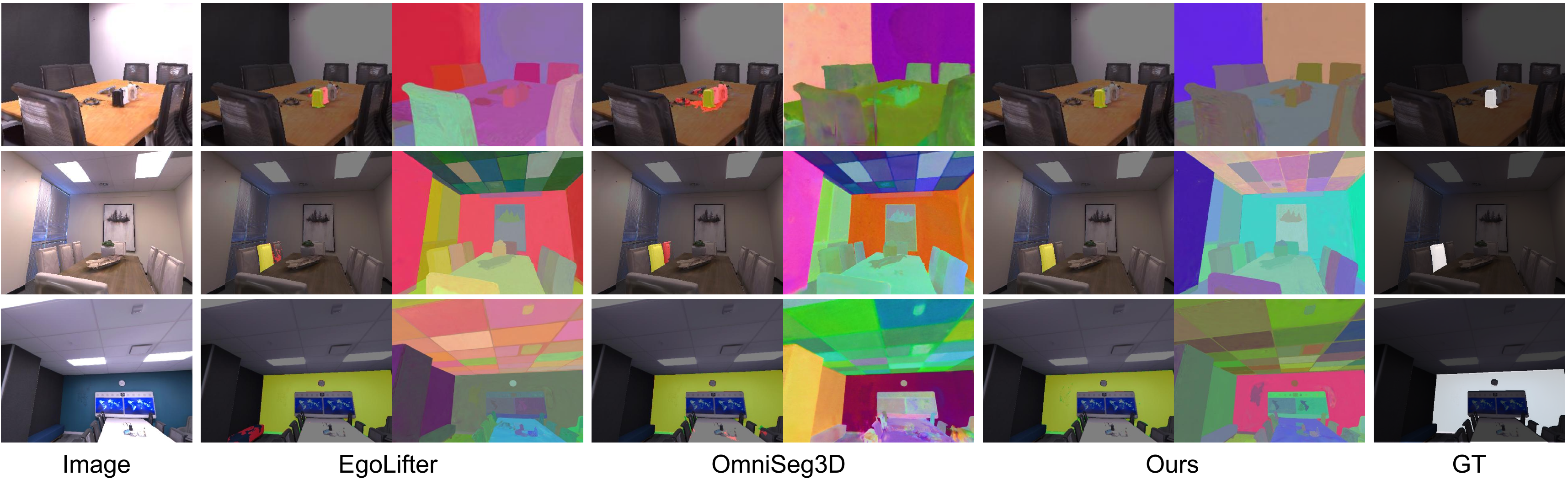}
    \caption{\textbf{Novel view synthesis of 2D segmentation on Replica.} The results of each baseline provide the segmentation on the left and the feature map from PCA on the right. The {\color{yellow}TP}, {\color{red}FP}, and {\color{green}FN} predictions are color-coded in the segmentation.}
    \label{fig:qual_replica}
\end{figure*}

\begin{table*}[ht]
\centering
\begin{minipage}[t]{0.71\linewidth}
    \centering
    \small
    \caption{\textbf{Novel view synthesis of 2D segmentation on the Replica dataset.}}
    \resizebox{\linewidth}{!}{
    \begin{tabular}{lccccccccc}
        \toprule
         Method & office0 & office1 & office2 & office3 & office4 & room0 & room1 & room2 & mean \\
         \midrule
         \rowcolor{gray!30} \multicolumn{10}{l}{\textbf{\textit{NeRF-based}}} \\ 
         MVSeg~\cite{mirzaei2023spin} & 31.4 & 40.4 & 30.4 & 30.5 & 25.4 & 31.1 & 40.7 & 29.2 & 32.4 \\
         SA3D~\cite{cen2023segment} &
         \cellcolor{gsecond}84.4 & 77.0 & \cellcolor{gthird}88.9 & \cellcolor{gthird}84.4 &
         \cellcolor{gbest}\textbf{82.6} & 77.6 & \cellcolor{gthird}79.8 & 89.2 &
         \cellcolor{gthird}83.0 \\
         OmniSeg3D~\cite{ying2024omniseg3d} &
         \cellcolor{gthird}83.9 & \cellcolor{gsecond}85.3 & \cellcolor{gsecond}89.0 &
         \cellcolor{gbest}\textbf{87.2} & 78.3 & \cellcolor{gsecond}83.0 &
         79.4 & \cellcolor{gsecond}88.9 & \cellcolor{gsecond}84.4 \\
         \midrule
         \rowcolor{gray!30} \multicolumn{10}{l}{\textbf{\textit{Gaussian Splatting-based}}} \\ 
         SA3D-GS~\cite{cen2023sags} & 82.1 & 72.7 & 81.3 & 83.2 & 65.7 &
         \cellcolor{gthird}79.9 & 79.0 & \cellcolor{gsecond}88.9 & 79.1 \\
         Egolifter~\cite{gu2025egolifter} & 82.9 & \cellcolor{gthird}78.4 & 85.1 & 84.1 & \cellcolor{gsecond}80.0 & 77.0 & \cellcolor{gbest}\textbf{86.4} & 84.3 & 82.1 \\
         Ours &
         \cellcolor{gbest}\textbf{84.7} & \cellcolor{gbest}\textbf{85.6} &
         \cellcolor{gbest}\textbf{90.2} & \cellcolor{gsecond}84.7 &
         \cellcolor{gthird}78.4 & \cellcolor{gbest}\textbf{86.0} &
         \cellcolor{gsecond}85.7 & \cellcolor{gbest}\textbf{89.0} &
         \cellcolor{gbest}\textbf{85.5} \\
        \bottomrule
    \end{tabular}
    }
    \label{tab:sam}
\end{minipage}
\hfill
\begin{minipage}[t]{0.28\linewidth}
    \centering
    \small
    \caption{\textbf{NVS 2D Seg. on NVOS.}}
    \resizebox{\linewidth}{!}{
    \begin{tabular}{lccc}
        \toprule
         Method & mIoU & mAcc & \\
         \midrule
         NVOS~\cite{ren-nvos} & 70.1 & 92.0 \\
         ISRF~\cite{isrfgoel2023} & 83.8 & 96.4 \\
         SA3D~\cite{cen2023segment} & 90.3 & 98.2 \\
         OmniSeg3D~\cite{ying2024omniseg3d} & 91.7 & 98.4 \\
         SA3D-GS~\cite{cen2023sags} & 90.9 & 98.3 \\
         FlashSplat~\cite{shen2025flashsplat} & 91.8 & \textbf{98.6} \\
         GaussianCut~\cite{jain2024gaussiancut} & \textbf{92.5} & 98.4 \\
         \midrule
         Ours & \textbf{92.5} & \textbf{98.6} \\
         \bottomrule
    \end{tabular}
    }
    \label{tab:nvos}
\end{minipage}
\end{table*}

\paragraph{Replica} 
We use the same testing split as SA3D~\cite{cen2023segment} and OmniSeg3D~\cite{ying2024omniseg3d}. Following previous work~\cite{ying2024omniseg3d,gu2025egolifter}, we obtain 2D instance segmentation via feature query after contrastive lifting.
We compare our method with both NeRF-based methods (\ie. MVSeg~\cite{mirzaei2023spin}, SA3D~\cite{cen2023segment} and OmniSeg3D~\cite{ying2024omniseg3d}) and Gaussian Splatting-based methods (\ie. SA3D-GS~\cite{cen2023sags} and Egolifter~\cite{gu2025egolifter}).

As shown in~\cref{tab:sam}, our method outperforms on the \ac{miou} metric all baselines including the \ac{nerf}-based methods for the first time in GS-based methods. From~\cref{fig:qual_replica}, Egolifter~\cite{gu2025egolifter} and OmniSeg3D~\cite{ying2024omniseg3d} often misclassify adjacent objects and fail to produce correct instance segmentations, highlighted in red as \textit{FP}. This can be reflected by their feature maps from PCA in the figure, as their inconsistent 2D masks often confuse the boundary features on nearby objects. In contrast, our improvements in consistent instance maps and ambiguous Gaussian reduction yield more precise novel view instance segmentation and feature maps, even for tiny objects on the table.

\begin{figure*}[!t]
    \centering
    \includegraphics[width=\linewidth]{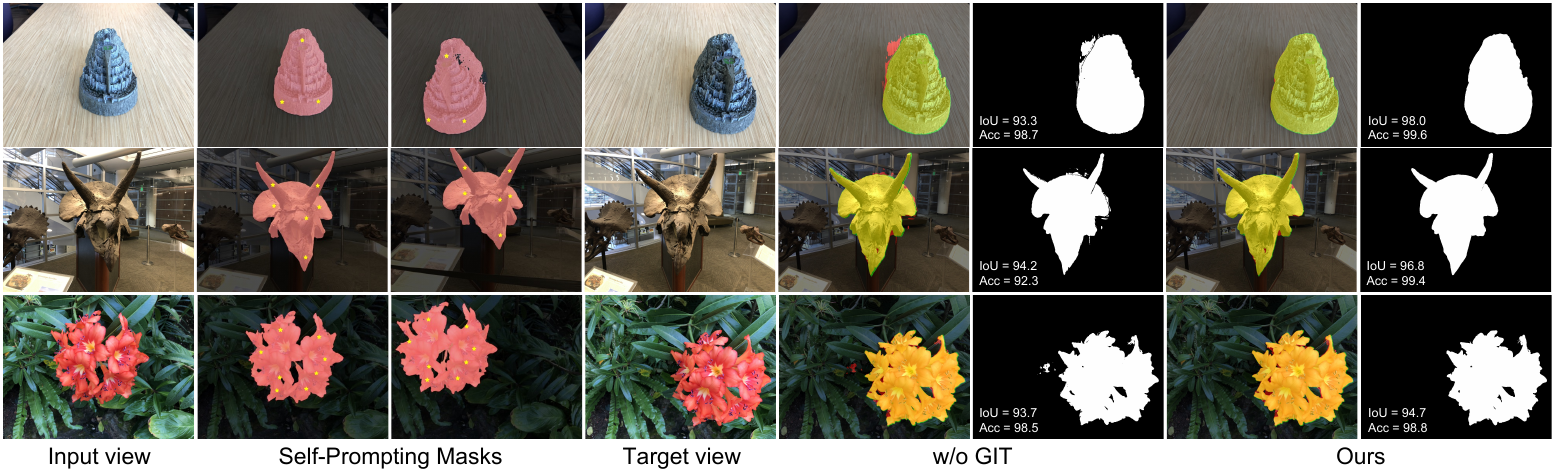}
    \caption{\textbf{Novel view synthesis of 2D segmentation on NVOS dataset.} The {\color{yellow}TP}, {\color{red}FP} and {\color{green}FN} predictions are color-coded in the segmentation. Our results with GIT produce fewer false positive predictions that are hard to observe from the input view.}
    \vspace{-2mm}
    \label{fig:qual_nvos}
\end{figure*}

\paragraph{NVOS} 
Following the common practice~\cite{cen2023segment,shen2025flashsplat} on the NVOS~\cite{ren-nvos} benchmark, we compare our method against state-of-the-art baselines, which typically perform better in the self-prompting setting. We also apply self-prompting to generate 2D instance segmentation on novel views.
\cref{tab:nvos} shows our method achieves the best performance on both \ac{miou} and mAcc metrics.
Furthermore, incorporating \ac{git}, which integrates information from other views through the weight matrix $\mW$, significantly reduces false positives (shown in red) compared to the pure self-prompting method, as illustrated in ~\cref{fig:qual_nvos}.

\begin{table}[ht]
    \centering
    \small
    \caption{\textbf{Ablation experiments.} ``Object mIoU'' and ``PSNR'' corresponds to metrics in 3D object extraction (\cref{sec:3d_obj_extract}) and ``NVS mIoU'' for the metric in \cref{sec:2d_ins}. ``SP.'' and ``Contr.'' denote the Self-Prompting and Contrastive Lifting respectively.}
    \resizebox{\linewidth}{!}{
    \begin{tabular}{ccccccc}
        \toprule
        \multirow{2}{*}{Rep.} &\multirow{2}{*}{Lifting} & Consistent & Amb. Gauss. & NVS & Object & \multirow{2}{*}{PSNR} \\
        & & 2D Mask & Control &  mIoU& mIoU & \\
        \midrule
        3DGS & Contr. & $\times$      & $\times$       & 87.2 & 60.9 & 20.8    \\
        3DGS & Contr. & $\checkmark$  & $\checkmark$   & 88.3 & 63.7 & 21.2    \\
        2DGS & Contr.       & $\times$      & $\times$       & 87.0 & 62.5 & 21.0    \\
        2DGS & Contr.       & $\times$      & $\checkmark$   & 87.3 & 70.1 & 22.4\\
        2DGS & Contr.       & $\checkmark$  & $\times$       & 89.2 & 63.6 & 21.2  \\
        2DGS & SP.          & $\checkmark$  & $\checkmark$   & 72.7 & 57.7 & 19.6  \\
        2DGS & Contr.       & $\checkmark$  & $\checkmark$   & 89.1 & 72.1 & 22.6 \\
        \bottomrule
    \end{tabular}
    }
    \label{tab:ablation}
\end{table}
\subsection{Ablation Studies}
We conduct extensive ablation experiments on Replica for our proposed components and analyze how the self-prompting and contrastive lifting work under different scenarios in \cref{tab:ablation}. We summarize the following observations:
\begin{enumerate}[nolistsep,noitemsep,leftmargin=*]
    \item The \acs{git}-guided adaptive density control effectively detects and mitigates the ambiguous Gaussians, which typically have low opacity and large scale, contributing minimally to the final rendering. By incorporating segmentation constraints as a form of regularization, our method significantly enhanced 3D object extraction, leading to a 13.4\% improvement in object \ac{miou} on \ac{2dgs} and a 4.6\% improvement on \ac{3dgs}. The gains are particularly notable on \ac{2dgs}, which may benefit more from our approach due to its inherently superior geometric quality.
    \item The consistent 2D mask generally improves both the 3D object extraction and novel view 2D segmentation as it's fundamental for 3D segmentation lifting. It even helps the object mIoU on top of our adaptive density control, highlighting how inconsistent masks may obscure features tied to each Gaussian.
    \item Self-prompting underperforms on the Replica dataset compared to contrastive lifting, which provides more coherent 3D representations but requires more optimization time. However, self-prompting is more flexible, especially for segmenting fine details (\eg, in the NVOS dataset), where contrastive lifting struggles with tiny or texture-like elements. This underscores the need for improved granularity and 3D consistency in future work.
\end{enumerate}

\subsection{Generalization and Robustness}
\begin{figure}[!t]
    \centering
    \includegraphics[width=\linewidth]{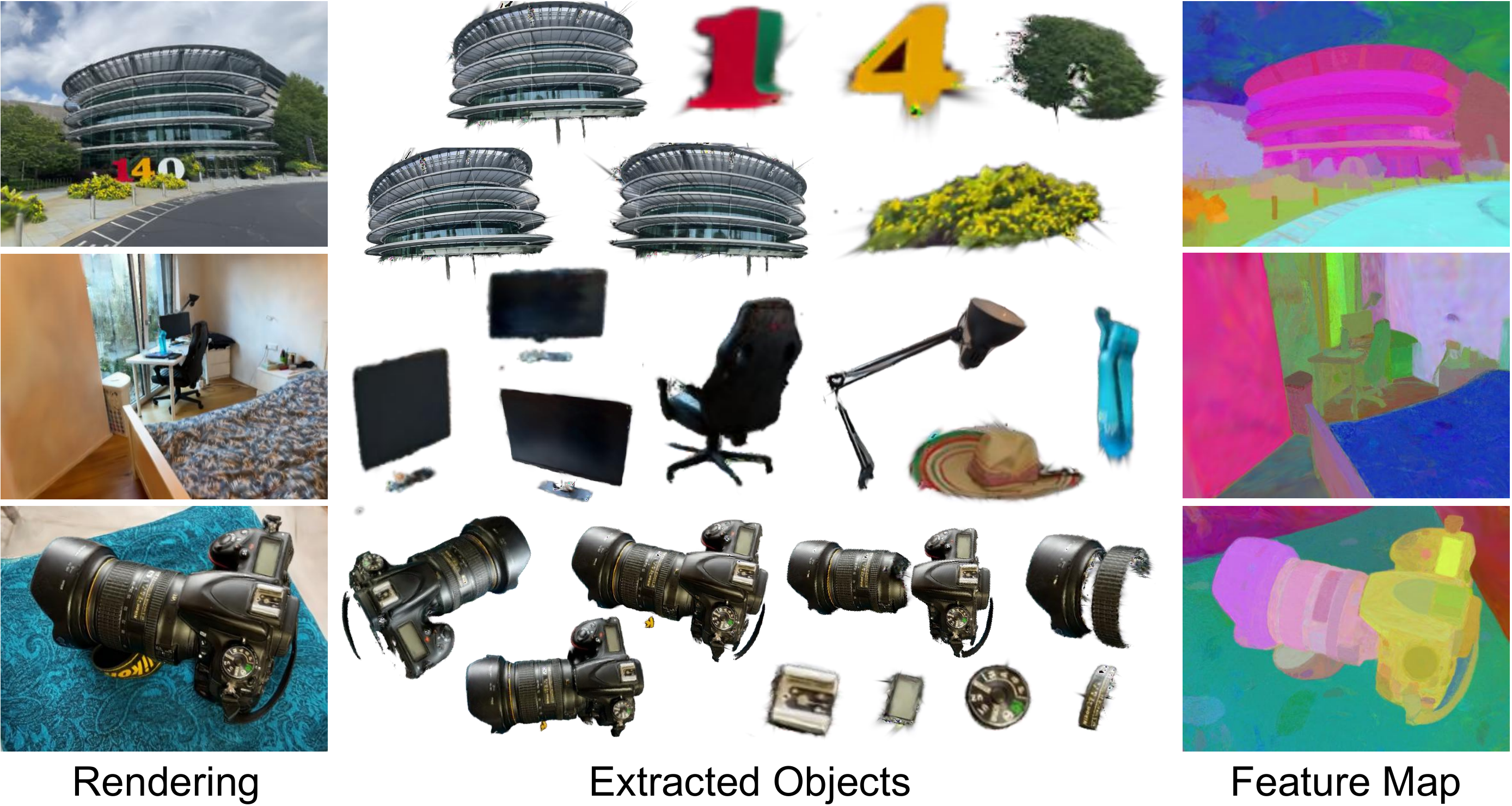}
    \caption{\textbf{Segmentation and object extraction on BlendedMVS~\cite{yao2020blendedmvs}, ScanNet++~\cite{yeshwanthliu2023scannetpp} and DL3DV-10K~\cite{ling2024dl3dv}.} Hierarchical object extraction is rendered on different views wth feature maps. }
    \label{fig:scene_editing}
    \vspace{-1mm}
\end{figure}

As demonstrated in \cref{fig:teaser,fig:scene_editing}, our method works robustly across diverse scenarios, including indoor/outdoor and bounded/unbounded, desktop settings, which only requires a query for the target object (or part) on a single reference view. Through 3D segmentation lifting via \ac{git}, we obtain Gaussian sets that can be rendered into novel views, produce clean and fine-grained feature maps, and enable hierarchical segmentation and 3D asset extraction. For example, the \textit{Camera} in \cref{fig:scene_editing} and \textit{Avengers} in \cref{fig:teaser} showcase our ability to extract arbitrary 3D objects, including specific parts.
These Gaussians facilitate diverse editing operations, \eg, translating, rotating, and recomposing objects by adjusting their locations and orientations as in \cref{fig:teaser}, or changing their style or colors by manipulating the Spherical Harmonics of the Gaussians.

\section{Conclusion}
This paper addresses the challenges in lifting 2D segmentation to Gaussians, \ie, the multiview inconsistent masks and ambiguous Gaussains. We propose \ac{git}, which assigns each Gaussian its instance probabilistic weights for all input views through a reverse-rasterization process. This enables majority-vote merging of fragmented 2D masks across different views and facilitates the identification of ambiguous Gaussians, which we then split or prune through guided density control; together they effectively enhance the quality of 3D segmentation lifting. Experiments on diverse datasets confirm our method’s validity and improved performance in both online and offline settings. Notably, our approach is also robust to in-the-wild video, supporting hierarchical object extraction and various downstream tasks in scene editing and object manipulation.

\clearpage
\small
\bibliographystyle{ieeenat_fullname}
\bibliography{main}

\clearpage






\appendix
\renewcommand{\thefigure}{S.\arabic{figure}}
\renewcommand{\thetable}{S.\arabic{table}}
\renewcommand{\theequation}{S.\arabic{equation}}
\maketitlesupplementary

\section{Failure Cases and Limitations}
\begin{figure}[ht!]
    \centering
    \includegraphics[width=\linewidth]{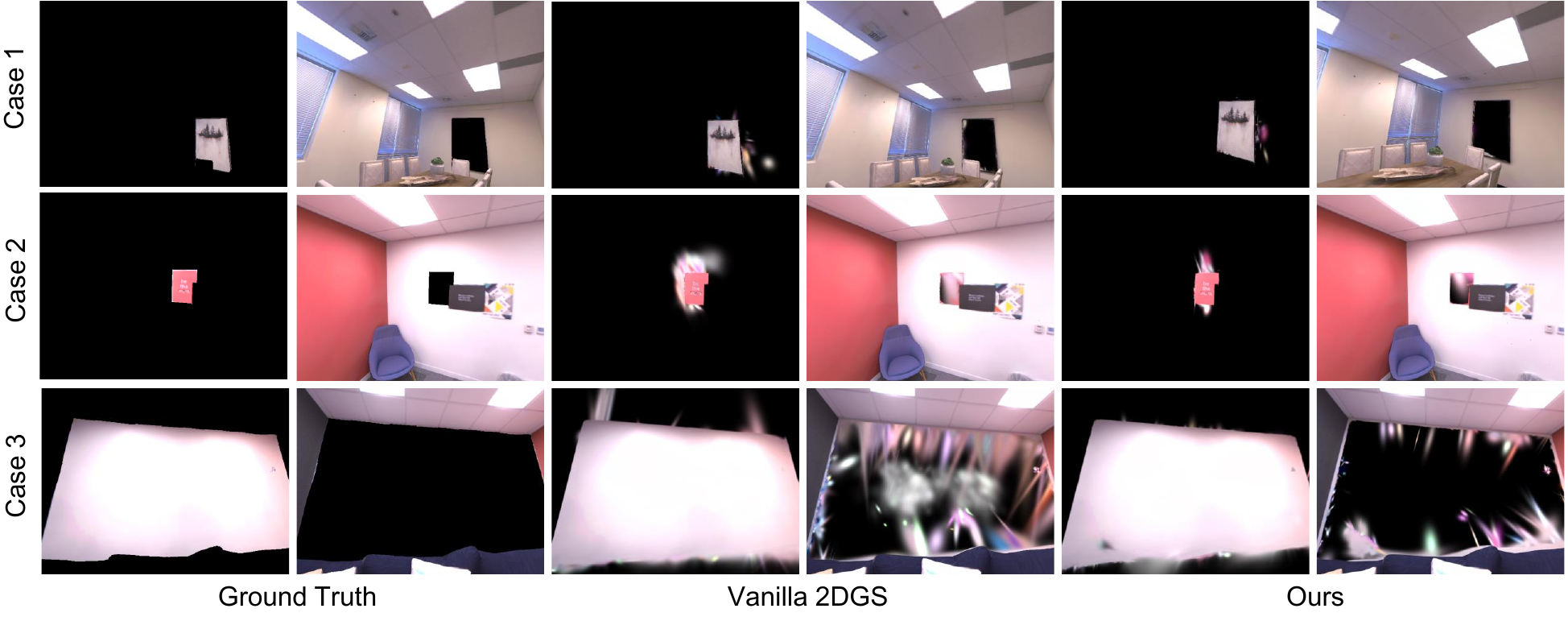}
    \caption{Failure cases on Replica.}
    \label{fig:failure_case}
\end{figure}

\paragraph{Failure Cases}
We visualize typical failure cases in \cref{fig:failure_case}. In Case 1, the artifact Gaussians are small and located far from the painting's surface. When traced from the forward view, it belongs to the painting, whereas from the lateral view, it belongs to the wall. In both views, this Gaussian only belongs to one object, albeit different ones. As a result, it is not considered ambiguous and we cannot eliminate them despite being clear artifacts. Case 2 and 3 share similar issues. Our \ac{git} is consistent with the rendering process and terminates when the accumulated opacity reaches 1. Therefore, if a Gaussian belongs to an object and is partially covered by a highly opaque surface, it cannot be flagged as ambiguous. These limitations degrade performance on fine-grained details, affecting hierarchical segmentation and object extraction.

\paragraph{Limitations}
We discuss two limitations that are valuable to address as future work.

First, although our motivation stems from leveraging the inherent consistency of Gaussians as a more explicit representation—ideally assigning one Gaussian per object or part—the approach still relies on a neural implicit framework with alpha blending for rendering. Consequently, there is a gap between true 3D surface geometry and the rendering-based association of Gaussians. As the level of granularity increases, especially for fine-grained details or texture-like patterns, the premise of consistent masks and the assumption about ambiguous Gaussians weaken. This also can be seen from Fig.~{\color{iccvblue}1} and Fig.~{\color{iccvblue}7} in the main paper, where hierarchical segmentation and object extraction exhibit more artifacts when the detail is refined to parts of a single body (\eg, Gundam) or object (\eg, camera). While our method is capable of robust segmentation and object extract on common objects under diverse scenarios, these findings underscore a recurring dilemma: while implicit representations such as \ac{nerf}~\cite{mildenhall2021nerf}, Gaussian Splatting~\cite{kerbl20233dgs,huang20242d}, and DMTet~\cite{shen2021deep} are relatively easy to optimize, fully explicit representations are often better suited for physics-based rendering and efficient simulation.

Second, although the \ac{git} operation is comparable in speed to forward rendering, merging inconsistent maps remains a relatively time-consuming step, as detailed in~\cref{subsec:scalability_of_patch_merge}. Future research could investigate more efficient data structures and algorithms to further mitigate training overhead.
Nevertheless, our Nonetheless, our \ac{git}-guided density control works as efficient as the original density control in original \ac{3dgs}~\cite{kerbl20233dgs}.

\begin{table*}[ht]
    \centering
    \small
    \vspace{-1em}
    \caption{The selected id lists used for 3D object extraction experiment in Replica.}
    \begin{tabular}{c|l}
        \toprule
         Scene & \multicolumn{1}{c}{ID list}  \\
         \midrule
         office0&3,4,7,9,11,12,14,15,16,17,19,21,22,23,29,30,32,34,35,36,37,40,44,48,49,57,58,61,66\\ 
         \midrule
         office1&3,8,9,11,12,13,14,17,23,24,29,30,31,32,34,35,37,43,45\\
         \midrule
         \multirow{2}{*}{office2}&0,2,3,4,6,8,9,12,13,14,17,23,27,34,38,39,46,49,51,54,57,58,59,63,65,68,69,70,72,73,74,75,77,78\\
                                  &80,84,85,86,90,92,93\\
         \midrule
         office3&1,2,8,11,12,15,18,21,22,25,29,32,33,42,51,54,55,56,60,61,70,82,85,86,88,86,97,101,102,103,110,111\\
         \midrule
         office4&3,4,5,6,9,13,16,18,20,23,31,34,47,48,49,51,52,56,60,61,62,65,69,70,71\\
         \midrule
        \multirow{2}{*}{room0} & 1,2,3,4,6,7,8,11,13,15,18,19,20,21,22,24,30,32,34,35,36,39,40,41,43,45,47,49,50,51,54,55,58,\\
                                & 61,63,64,68,69,70,71,72,73,74,75,78,79,83,85,86,87,90,92 \\
         \midrule
         room1& 3,4,6,7,8,9,11,12,13,15,17,18,19,21,22,23,24,27,30,32,33,35,37,39,40,43,45,46,48,50,51,52,53,54\\
         \midrule
         \multirow{2}{*}{room2}&2,4,5,10,14,15,17,18,19,20,22,24,26,27,28,29,31,32,34,36,38,39,40,42,44,46,47,48,49,52,54,55,56,\\
         &57,58,59,61\\
          \bottomrule
    \end{tabular}
    \label{tab:selected_object}
\end{table*}

\section{Method Details}
\label{sec:method_details}
\subsection{Scalability of GIT}
The weight matrix \scalebox{0.8}{$\mathbf{W} \in \mathbb{R}^{N \times T\times L}$}, where $N$ is the number of Gaussians, $T$ the maximum number of patches in any single view, and $L$ the number of views, is conceptually defined but not explicitly used during implementation. In practice, we accelerate GIT by first computing a temporary $N \times T$ matrix sequentially for each view, which is then reduced to an $N$-dim vector storing the patch ID with the highest probability. The vectors for all $L$ views are combined into a $N \times L$ matrix for patch merging and GS refinement. Thus, our method only maintains a temporary $N \times T$ matrix and a cumulative $N \times L$ matrix during training, allowing us to handle scenes with a large number of objects efficiently.
\subsection{Scalability of patch merging}
\label{subsec:scalability_of_patch_merge}
This step involves computing patch similarity by tracing all relevant Gaussians and checking whether they belong to the same patch based on majority votes across all views. The theoretical upper bound on complexity is $O(N^2\cdot T^2)$.
While in implementation, we avoid computing similarities for all primitive-level patch pairs within a view based on SAM’s hierarchical information: only patches that fall within the same region of coarser-level masks are considered for merging.
Furthermore, patch similarity is computed only over co-visible views shared by the two Gaussian sets, rather than across all views. Combined with our \acs{git} acceleration, they ensure efficiency in large-scale, multi-object scenes.
\begin{figure}[h]
    \centering
    \includegraphics[width=\linewidth]{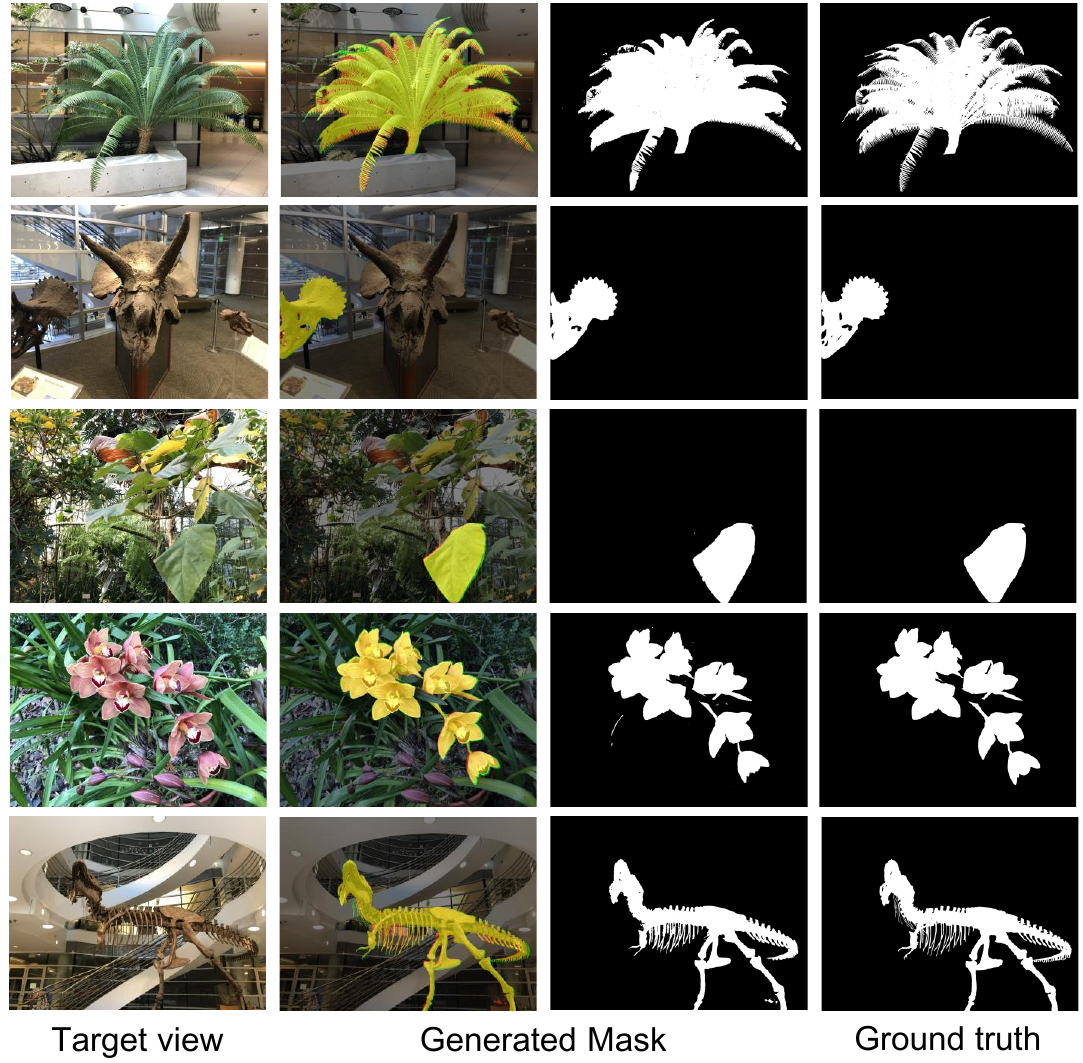}
    \vspace{-1.5em}
    \caption{Remaining visualization results on NVOS.}
    \label{fig:more_visual_nvos}
\end{figure}
\section{Method Comparison}

Previous work~\cite{gu2025egolifter, ying2024omniseg3d} also proposes an overlay mask solution to alleviate inconsistencies in multi-view masks generated by SAM~\cite{kirillov2023segment}. Here, we provide a detailed discussion comparing our method with these approaches.

\paragraph{EgoLifter}
EgoLifter~\cite{gu2025egolifter} discards information about overlapping regions and simply overlays all masks in image space to obtain a one-hot segmentation for each pixel. However, due to the randomness of the topmost mask across different views, EgoLifter's solution fails to resolve inconsistencies in the overlapping areas.

\paragraph{Omniseg3D}
Similarly to our approach, OmniSeg3D~\cite{ying2024omniseg3d} also divides the 2D image into disjoint patches by overlapping the SAM masks. it models the hierarchical structure within these patches by measuring correlations, characterized by the number of masks that contain the corresponding masks. During contrastive lifting, it enforces this hierarchy through an explicit ordering regularization. While the concept is promising, inconsistent SAM predictions can produce ambiguous features across different hierarchical levels, especially in fine-grained cases. This issue is evident in the qualitative comparisons in the main paper.

\paragraph{GarField}
GarField~\cite{kim2024garfield} optimizes a scale-conditioned affinity field in an attempt to alleviate inconsistencies in multi-view masks. However, its scale conditioning is overly sensitive and operates as a black-box, making it challenging to use during the inference phase.

\paragraph{Gaga}
While both our method and Gaga~\cite{lyu2024gaga} leverage the mask-Gaussian relationship, Gaga focuses on 3D lifting with mask ID association, whereas our method addresses inconsistent SAM masks and refines Gaussians, benefiting general 3D segmentation lifting methods. Gaga traces via Gaussian center projection, which struggles with occlusion, while our reverse rasterization offers more precise, view-consistent alignment. Moreover, Gaga uses a sequential 3D memory bank, where assignments are fixed once added, limiting effective cross-view aggregation. In contrast, our majority voting across all views enables more robust and consistent segmentation. 

\paragraph{Ours}
We obtain a single instance map by overlapping all masks and treating each overlapping region as a disjoint instance patch. We then perform consistent masking and \ac{git}-guided density control on these primitive-level patches, providing clear guidance during contrastive lifting. This ensures that features are distinctly grouped or separated, leading to sharper boundaries, clearer feature maps, and flexible segmentation or object extraction across various levels of granularity in our experiments.

\begin{table*}[t!]
    \centering
    \small
    \vspace{-0.5em}
    \caption{Quantitative results of 3D object extraction on Replica across all scenes.}
    \vspace{-0.8em}
    \resizebox{\linewidth}{!}{
    \begin{tabular}{l|ccccccccc|c}
        \toprule
         Method & office0 & office1 & office2 & office3 & office4 & room0 & room1 & room2 & \textit{avg.} mIoU &\textit{avg.} PSNR \\
         \midrule
         Gaussain Grouping~\cite{gaussian_grouping}&23.7&45.9&25.5&30.6&30.2&22.5&38.5&20.1&29.6&13.4\\
         FlashSplat~\cite{shen2025flashsplat}      &47.5&45.9&39.6&36.9&27.5&40.6&39.8&36.6&39.3&16.9\\
         Egolifter~\cite{gu2025egolifter}          &67.4&59.6&48.9&54.8&59.4&50.7&53.7&50.1&55.6&20.1\\
         Gaga~\cite{lyu2024gaga}                   &45.1&47.8&37.8&37.2&40.4&39.3&36.9&44.5&41.1&17.6\\
         Ours&\textbf{80.7}&\textbf{76.0}&\textbf{66.1}&\textbf{69.8}&\textbf{71.7}&\textbf{67.0}&\textbf{72.0}&\textbf{73.4}&\textbf{72.1}& \textbf{22.6}\\
        \bottomrule
    \end{tabular}
    }
    \label{tab:gaussian_seg_iou}
\end{table*}

\begin{table*}[t!]
    \centering
    \small
    \vspace{-0.5em}
    \caption{ The detailed PSNR of 3D object extraction on Replica across all scenes.}
    \vspace{-0.8em}
    \resizebox{\linewidth}{!}{
    \begin{tabular}{l|cccccccc|c}
        \toprule
         Method & office0 & office1 & office2 & office3 & office4 & room0 & room1 & room2  &\textit{avg.} PSNR \\
         \midrule
         Gaussain Grouping~\cite{gaussian_grouping}&16.0&22.8&10.5&11.2&12.3&10.3&12.7&11.7&13.4\\
         FlashSplat~\cite{shen2025flashsplat}      &21.2&24.5&14.4&14.0&14.9&14.8&15.2&16.2&16.9\\
         Egolifter~\cite{gu2025egolifter}          &26.5&29.1&16.3&16.0&19.5&17.0&17.4&19.1&20.1\\
         Gaga~\cite{lyu2024gaga}                   &22.4&27.0&13.5&13.9&16.7&14.5&14.7&18.0&17.6\\
         Ours&\textbf{28.1}&\textbf{29.9}&\textbf{18.9}&\textbf{20.0}&\textbf{22.0}&\textbf{19.4}&\textbf{20.3}&\textbf{22.3}& \textbf{22.6}\\
        \bottomrule
    \end{tabular}
    }
    \label{tab:gaussian_seg_psnr}
\end{table*}

\begin{figure*}[t!]
    \centering
    \includegraphics[width=\linewidth]{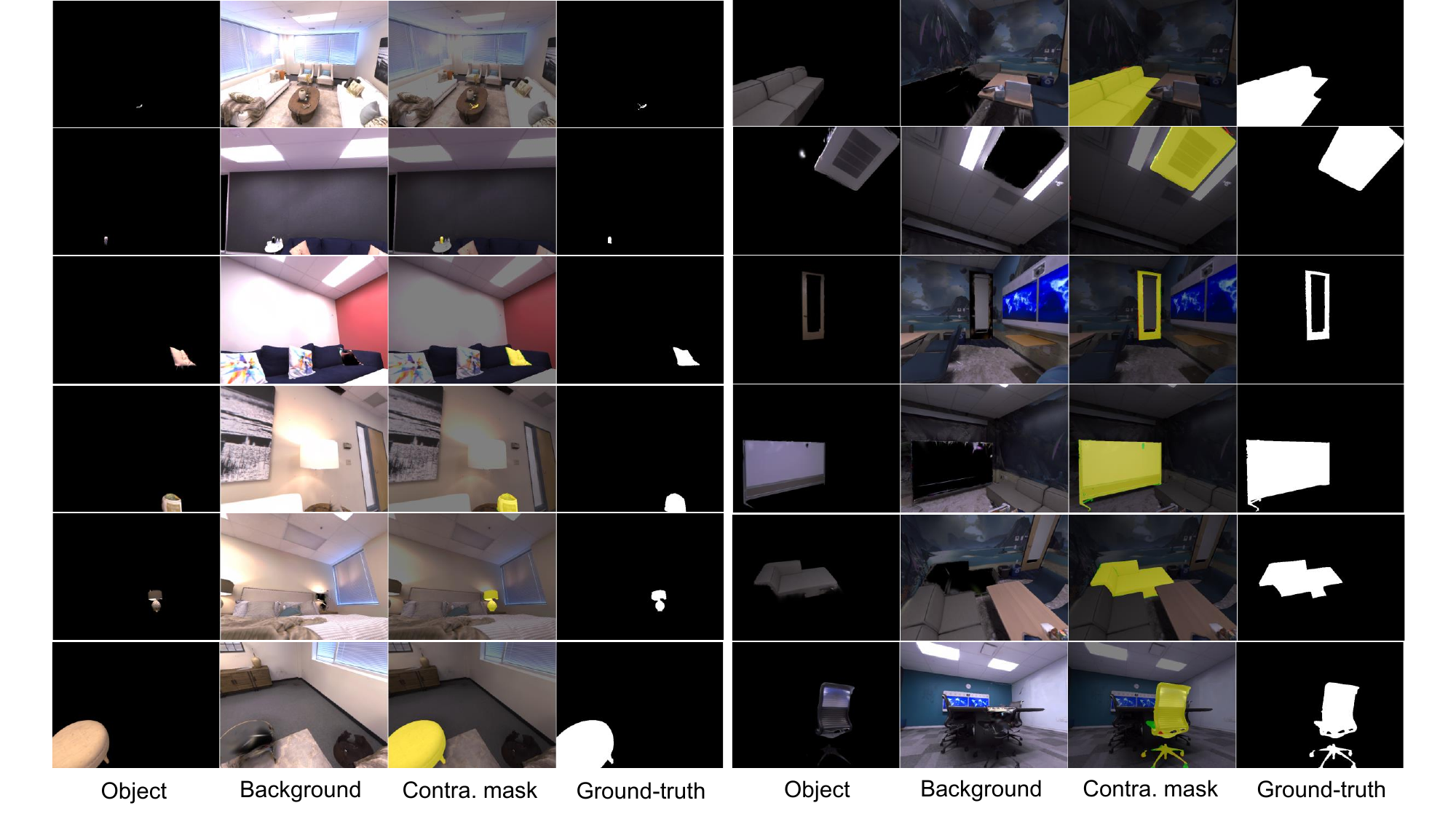}
    \caption{More visualization results on Replica.}
    \label{fig:more_visual_replica}
\end{figure*}

\section{Experiments: 3D Object Extraction}

\subsection{Evaluation Details}
We render the extracted 3D object into novel views to generate its corresponding RGB image. The regions where the RGB values exceed zero are utilized as a mask to compute the IoU with the ground-truth mask. For PSNR calculation, we restrict the computation to a specific region defined by the bounding box of the ground-truth mask, expanded outward by 10 pixels. This ensures that the evaluation focuses on the relevant object regions while reducing the influence of background areas. 

\subsection{Testing Split}
We select the majority of instances from the Replica dataset, excluding floors, ceilings, excessively large walls, and low-quality objects. The full list is provided in \cref{tab:selected_object}. To more accurately evaluate the quality of the extracted 3D objects, we filter out test views where the objects are occluded.

\section{More Experiment Results}
We provide detailed results on Replica in \cref{tab:gaussian_seg_iou} \cref{tab:gaussian_seg_psnr}. Additionally, we present further qualitative results on Replica in \cref{fig:more_visual_replica} and on the remaining NVOS scenes in \cref{fig:more_visual_nvos}.

\end{document}